# Human Activity Recognition Using LSTM-RNN Deep Neural Network Architecture

Schalk Wilhelm Pienaar[1], Reza Malekian[1,2], Senior Member, IEEE,
[1]Department of Electrical, Electronic and Computer Engineering, University of Pretoria, Pretoria,0002, South Africa

[2]Department of Computer Science and Media Technology, Malmö University, Malmö, 20506, Sweden/ Internet of Things and People Research Center, Malmö University, Malmö, 20506, Sweden
Reza.malekian@ieee.org

*Abstract*—Using raw sensor data to model and train networks for Human Activity Recognition can be used in many different applications, from fitness tracking to safety monitoring applications. These models can be easily extended to be trained with different data sources for increased accuracies or an extension of classifications for different prediction classes. This paper goes into the discussion on the available dataset provided by WISDM and the unique features of each class for the different axes. Furthermore, the design of a Long Short Term Memory (LSTM) architecture model is outlined for the application of human activity recognition. An accuracy of above 94% and a loss of less than 30% has been reached in the first 500 epochs of training.

## I. Introduction

With recent technology advances in mobile devices such as smart phones, many sensors, part of these devices, have also improved greatly over the past few years. These include motion sensors, light sensors, GPS, image-sensors (camera/s), sound sensors (microphone), etc. With the computing performance increases in these devices, the ability to perform tasks, that would have been considered too processing-intensive in real-time a few years back, have been made possible, without requiring the data to be transmitted to a server for processing. Due to these advancements, there are exciting new areas of development that are opened up, which can assist in many areas of our daily lives. The goal of this paper is to investigate and build a solution to identify the activities by making use of an open-source dataset, released by the Wireless Sensor Data Mining (WISDM) Lab [1]. The dataset includes labels for each of the six activities, each including the x, y and z axis values for the tri-axial accelerometer during the labelled activities. The activities available in the dataset includes standing, sitting, walking, jogging, ascending and descending stairs.

Machine-learning can be performed with this data using various methods; this paper will analyse Recurrent Neural Networks (RNN) with Long Short-Term Memory (LSTM) cells. LSTM has feedback connections, providing it with the ability to compute everything that a Turing machine can perform [2]. It can also process both single data points, such as images, and sequences of data, such as speech, video, human activity, etc. LSTM units each consist of a cell, and three regulators/gates, namely an input and output gate, and a forget gate. Each cell remembers values of arbitrary time-intervals, while the gates regulate the information flow both from and to the cells. These networks are ideal for classification, processing and predicting time-series based data. LSTMs is proven to excel in learning, processing and classifying such types of data.

In traditional RNNs, where the network is trained via backpropagation through time, an issue known as exploding and vanishing gradients, are present [3], causing an exploding gradient problem. This is when large error gradients start accumulating, resulting in significant changes to the neural network model during training, which in effect prevents a model from training with the available data, and causes the trained model to be unstable [4]. Vanishing gradients is when the gradients of loss functions become too small (approaches zero), then the network becomes increasingly hard to train, as the weights and biases of the initial layers are not updated effectively with the training sessions [5]. LSTM has been designed to overcome these issues that arise in RNN architectures. LSTM and RNNs are very similar, with the difference being that hidden layers in LSTMs contain memory blocks with cells instead of non-linear units, which can store information over long time-spans. In other words, traditional RNN cells have a single internal layer that acts on the current state ($h_{t-1}$) and input ($x_t$), whereas an LSTM cell contains three such layers [6].

## II. Model Training

### A. Raw Data Activity analysis

This study considers six activities, namely jogging, sitting, standing, walking, upstairs and downstairs. These activities are available in raw data, provided by WISDM, and is ideal to use as a base for our research. Most of these activities also involves repetitive motion making it easier to train and recognize. The activities have been recorded using a tri-axial accelerometer. The x-axis represents the horizontal motion, which is towards the left and right sides of the phone. This is used to capture the horizontal motion of the leg. The y-axis detects the vertical motion which is the direction to the top and bottom of the phone, while the z-axis represents the motion into and out of the screen, which can be used to capture the leg's backword and forward motion.

In general, all the considered activities, excluding sitting and standing, can be described by the time between acceleration peaks and the magnitude of these values.

Figure 1 to Figure 3, below, shows the training accelerometer data for the x, y and z axes for the jogging class. The x-axis clearly distinguishes itself from the other classes x-axis data, with the values fluctuating between -10 and 10, unlike the rest of the classes. The duration between peaks, in general, is also shorter than activities such as walking.

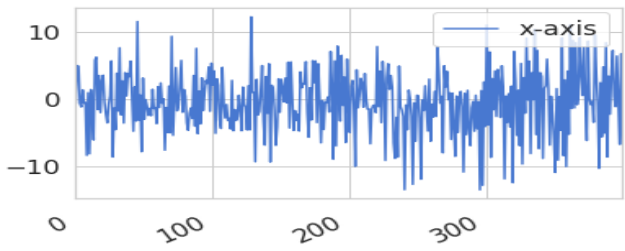

Figure 1: jogging x-axis

The y-axis data shown in **Error! Reference source not found.** is quite distinguishable compared to other classes, except for the comparison with the walking class, where both fluctuate around 0 to 20. Most of the y-axis data, where the user is in an upright position, has a value of about 9.8 m/s$^2$ as a result of the earth's gravitational pull.

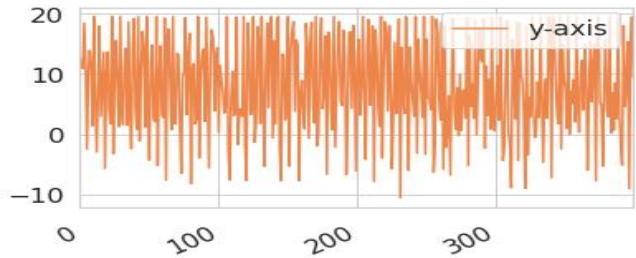

Figure 2: jogging y-axis

The z-axis plot for jogging, in Figure 3, is also quite similar to the z-axis plot for walking, but with the values fluctuating more greatly into the negative range, which is expected with the type of activity, where the user uses more force in their motion.

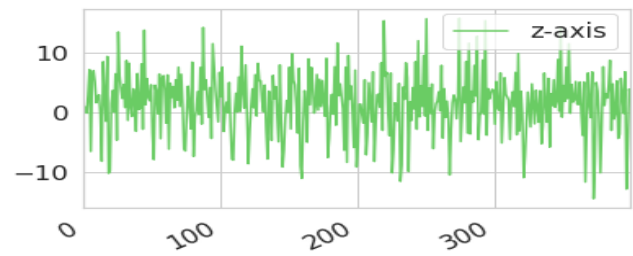

Figure 3: jogging z-axis

For the sitting classes' x-plot, the plot remains almost constantly stable, similar to the standing classes' plot, however, the x-axis value for sitting is at 3 as opposed to almost zero for the sitting class. It should be noted that the differences between the sitting and standing activities is the relative magnitudes, due to the differences in the device orientation, and because of this, the network would be able to be trained to make a proper distinction.

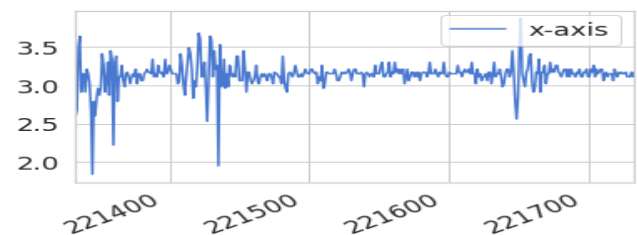

Figure 4: sitting x-axis

In Figure 5, the y-axis data is plotted for sitting, which is barely distinguishable to that of the standing graph, except that it slightly more stable.

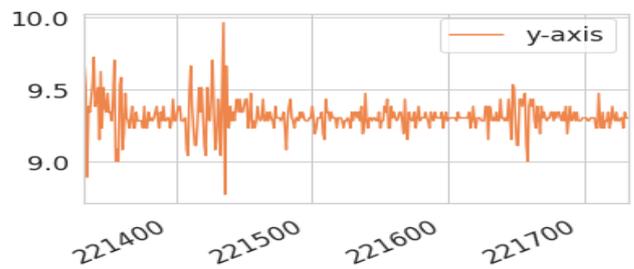

Figure 5: sitting y-axis

Once again, Figure 6 indicates that comparing standing with sitting proves to be quite difficult to distinguish, since in both states, the user does not exert a lot of motion on the mobile device.

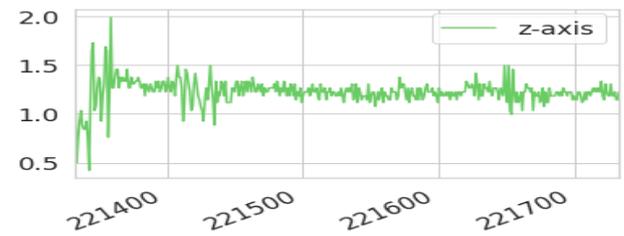

Figure 6: sitting z-axis

Figure 7 - Figure 9 represent the plotted data for the x, y and z axes, respectively. The x-axis is quite stable at approximately zero, due to almost no motion. The y-axis is quite stable at about 9.8m/s$^2$, due to a constant gravitational pull, and the z-axis is very similar to the walking graph, but without the constant peaks that are caused by the device moving back and forth in a direction perpendicular to the screen.

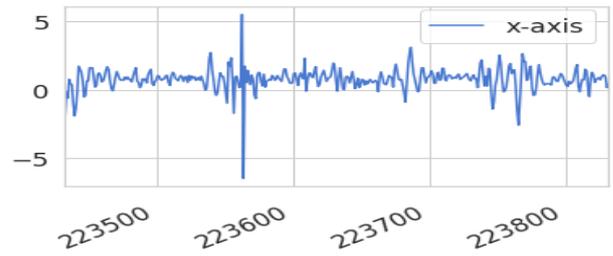

Figure 7: standing x-axis

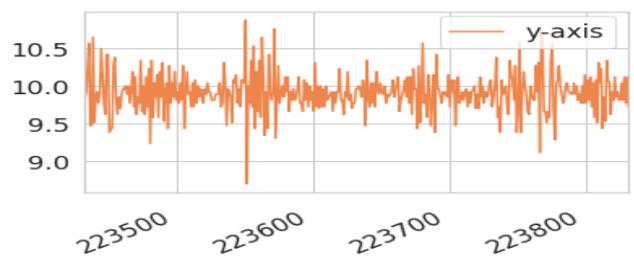

Figure 8: standing y-axis

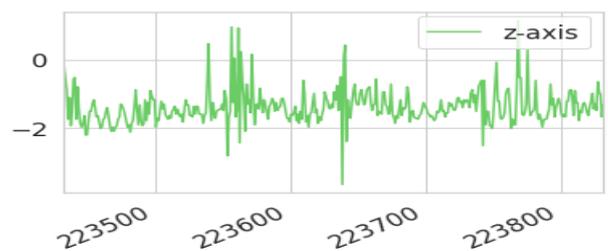

Figure 9: standing z-axis

Figure 10 - Figure 12 is a representation of the data available for the walking data class, in the x, y and z-axis, respectively. The peaks for each of the three axes are spaced more or less the same, but each with its own intensity and range.

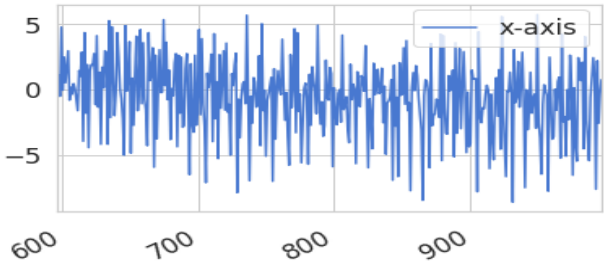

**Figure 10: walking x-axis**

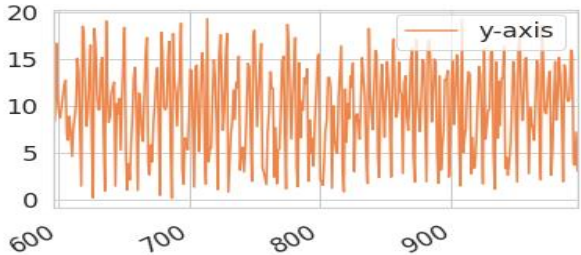

**Figure 11: walking y-axis**

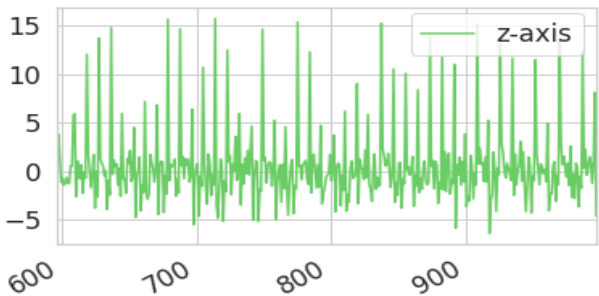

**Figure 12: walking z-axis**

Figure 13 - Figure 15 represents the graphs for the downstairs data. The x-axis shows a series of regular peaks fluctuating around zero. The y-axis values spikes to about 15 with each step being taken, and the z-axis is less aggressive than that of jogging, but mostly similar.

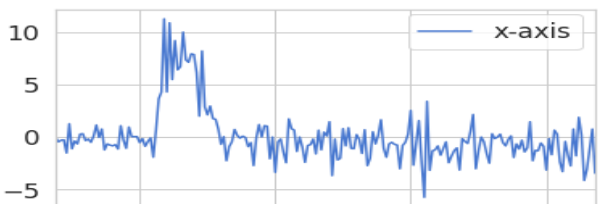

**Figure 13: Downstairs x-axis**

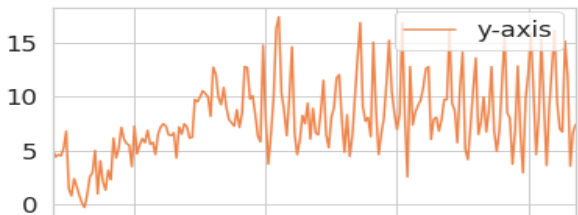

**Figure 14: downstairs y-axis**

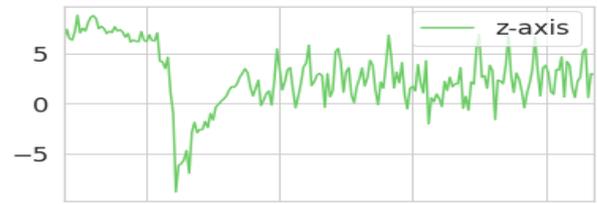

**Figure 15: downstairs z-axis**

### B. Feature Generation and Data Transformation

In order to have many training batches at hand, it is necessary to split each class into a set number of data points, and to store this into segments. An optimal size is about 200 points, which works out to be about 10 seconds worth of data at a time with a step size of 20. The segments can then be fed into our network for training, where the network will learn itself what features belong to what class, based on the labels provided with the training data. When the real-time data is fed into the trained network, it will then be able to make predictions.

The following method uses a windowing method to extract 200 data points from the dataset to create segments and labels for our model training [7].

```
1. segments = []
2. labels = []
3. for i in range(0, len(df) - TIME_STEPS, STEP):
4.     xs = np.asarray(df['x-axis'][i: i + TIME_STEPS])
5.     ys = np.asarray(df['y-axis'][i: i + TIME_STEPS])
6.     zs = np.asarray(df['z-axis'][i: i + TIME_STEPS])
7.     label = stats.mode(df['activity'][i: i + TIME_STEPS])[0][0]
8.     segments.append([xs, ys, zs])
9.     labels.append(label)
10. segments_transformed = np.asarray(segments, dtype= np.float32).reshape(-1, TIME_STEPS, FEATURES)
11. labels = np.asarray(pd.get_dummies(labels), dtype = np.float32)
```

### C. Data Processing

From our dataset, it is necessary to split the data into training, testing and validation sets for modelling purposes. A competing concern is that, with a smaller training set, the parameter estimates have more variance. However, with a smaller test set, the performance statistics will have a greater variance. It is important that neither of the variances are too high. With large datasets, whether it is a 70:30 or a 90:10 split, this does not influenced, since both splits should be able to give satisfactory variances in the estimates. The reason for having validation data is to help evaluate the quality of the model, to avoid over- and under-fitting, and to assist in selecting the model that would perform the best on unseen data. The steps required for data separation are therefore:

- split data into 80:20, training:testing, for instance
- split the remaining training data into 80:20, training:validation
- Sub-sample a couple of random selections of the training data and train the model with this and record the performance on the validation set.
- Perform the sequence with different percentages of training and testing data to get the optimal value.

```
1. train_test_split = np.random.rand(len(segments_transformed)) < 0.80
2. xTrain = segments_transformed[train_test_split]
3. yTrain = labels[train_test_split]
4. xTest = segments_transformed[~train_test_split]
5. yTest = labels[~train_test_split]
```

*D. Modeling*

Defining the network inputs is the first requirement when building a neural network. For this, TensorFlow placeholders are defined for the data sequences and their labels [8].

```
1.  INPUT = tf.placeholder(tf.float32, [None, TIME_STEPS, FEATURES], name="input")
2. OUTPUT = tf.placeholder(tf.float32, [None, CLASSES])
```

Placeholders are simply variables (places in memory) for the data that will be fed into our network during training. It makes it possible to create our operations and build a computation graph. The placeholders are then used to allow data to be fed into this graphs [9].

It is then required to set up the LSTM layers. We will stack 3 LSTM layers, each with 64 units.

```
12. def LSTM_model(inputs):
13.     hidden_bias = tf.Variable(tf.random_normal([HIDDEN_UNITS], mean=1.0))
14.     output_bias = tf.Variable(tf.random_normal([CLASSES]))
15.     hidden_weights = tf.Variable(tf.random_normal([FEATURES, HIDDEN_UNITS]))
16.     output_weights = tf.Variable(tf.random_normal([HIDDEN_UNITS, CLASSES]))
17.     INPUT = tf.transpose(inputs, [1, 0, 2])
18.     INPUT = tf.reshape(INPUT, [-1, FEATURES])
19.     hdn = tf.nn.relu(tf.matmul(INPUT, hidden_weights) + hidden_bias)
20.     hdn = tf.split(hdn, TIME_STEPS, 0)
21.     # 3 LSTM layer stacking
22.     lstm_layers = [tf.nn.rnn_cell.LSTMCell(HIDDEN_UNITS, forget_bias=1.0) for _ in range(3)]
23.     lstm_layers = tf.nn.rnn_cell.MultiRNNCell(lstm_layers)
24.
25.     outputs, _ = tf.contrib.rnn.static_rnn(lstm_layers, hdn, dtype=tf.float32)
26.
27.     return tf.matmul(outputs[-1], output_weights) + output_bias
```

*E. Training*

Now that all the pieces have been set up of the computation graph, it is possible to build the graph and train the network. Once the network has been trained, it can be exported to be used with a mobile application.

Choosing a learning rate can prove to be quite tedious and is different for each optimiser; a learning rate that is too low will take either very long to progress or simply fail, whereas a learning rate that is too high would cause instability and have difficulty converging. Learning rates can affect the training time by as much as an order of magnitude. It should also be noted that training time increases linearly with the model size and that the choice of hyper-parameters (i.e. learning-rate and dropout-rate) remains valid despite the model being scaled linearly, making it possible to fine-tune these parameters on a scaled down version of the network.

Cost, otherwise known as loss, is calculated by comparing the prediction with the ground-truth. The Adam optimiser is used together with the cost to optimise it. The Adam optimization algorithm aims to find the optimum weights, minimise errors, and maximise accuracy [10]. It is easy to implement, computationally efficient, and appropriate for large datasets and non-stationary classifications [10]. L2 Regularization is implemented, which is an important technique in machine-learning that prevents overfitting by adding a regularization term (i.e., it introduces additional information) which in effect allows the model to make correct predictions on un-seen data. L1 regularization could also have been used, which is very similar to L2, except that L2 is the sum of the squares of weights instead of the sum of weights, and it is more computationally efficient, together with some other minor differences [11].

**Building the graph:**

```
3. tf.reset_default_graph()
4. INPUT = tf.placeholder(tf.float32, [None, TIME_STEPS, FEATURES], name="input")
5. LABELS = tf.placeholder(tf.float32, [None, CLASSES])
6. prediction = LSTM_model(INPUT)
7. predict_softmax = tf.nn.softmax(prediction, name="y_")
8.
9. l2_regularization = 0.0015 * sum(tf.nn.l2_loss(trainable_var) for trainable_var in tf.trainable_variables())
10.
11. cost = tf.reduce_mean(tf.nn.softmax_cross_entropy_with_logits(logits = prediction, labels = LABELS)) + l2_regularization
12.
13. optimizer = tf.train.AdamOptimizer(learning_rate=0.0025).minimize(cost)
14. correct_prediction = tf.equal(tf.argmax(predict_softmax, 1), tf.argmax(LABELS, 1))
15. accuracy = tf.reduce_mean(tf.cast(correct_prediction, dtype=tf.float32))
```

**Training the network:**

```
1. saver = tf.train.Saver()
2.
3. with tf.Session() as session:
4.     session.run(tf.global_variables_initializer())
5.     train_count = len(xTrain)
6.
7.     for i in range(NO_EPOCHS):
8.         for start, end in zip(range(0, train_count, BATCH_SIZE),
9.                               range(BATCH_SIZE, train_count + 1, BATCH_SIZE)):
10.             session.run(optimizer, feed_dict={INPUT: xTrain[start:end], LABELS: yTrain[start:end]})
11.
12.         _, train_accuracy, train_loss = session.run([predict_softmax, accuracy, loss], feed_dict={INPUT: xTrain, LABELS: yTrain})
13.         _, test_accuracy, test_loss = session.run([predict_softmax, accuracy, loss], feed_dict={INPUT: xTest, LABELS: yTest})
14.
15.         if i != 1 and i % 10 != 0:
16.             continue
17.
18.         print(f'epoch: {i}: loss: {test_loss}, test accuracy: {test_accuracy}')
19.
20.     predictions, final_accuracy, final_loss = session.run([predict_softmax, accuracy, loss], feed_dict={INPUT: xTest, LABELS: yTest})
21.
22.     print('-----------------')
23.     print(f'final results: loss: {final_loss}, accuracy: {final_accuracy}')
```

## III. RESULTS

Figure 16 shows the trained model's prediction performance in the form of a confusion matrix. The greatest confusion lies in the prediction of upstairs and downstairs that is confusion with walking. The interpretation of up and downstairs as walking may not be necessarily incorrect, but it is a misclassification considering the available classes. There is also some confusion between sitting and standing, although the prediction is surprisingly good.

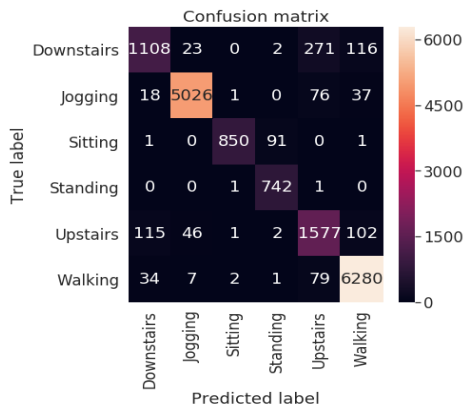

**Figure 16: Confusion matrix**

Figure 17 shows the progression of the model training in terms of accuracy and loss over the number of Epochs. The number of Epochs is a measurement that defines the total iterations, where each iteration consists of all training vectors in the dataset being passed forward and backwords through the neural network and is used for updating the weights. For sequential training, the weights are all updated with each training vector that is passed through, unlike the case for batch training, where the entire batch needs to be passed through the algorithm, before updating the weights [12]. Batch training was used with a batch size of 1024. From the figure, it can be seen that the test and train accuracy [13] reached about 90% and the train loss went down to about 40% with the first 200 epochs. The loss is used to optimise the machine learning algorithm [14,15], whereas the accuracy [16,17, 18] is used to measure the performance of the algorithm.

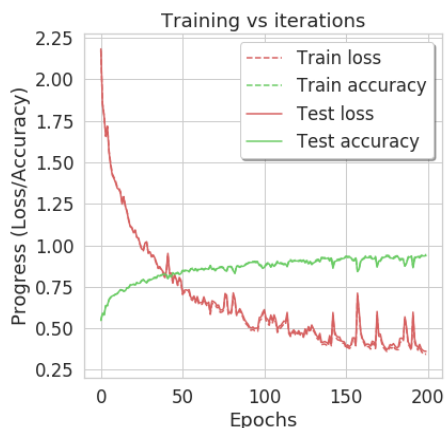

**Figure 17: training progress vs epochs**

## IV. CONCLUSION

In this paper, a human activity recognition design has been proposed using an LSTM-RNN deep neural architecture. The network is able to learn all six classes successfully and efficiently with only a few hundred epochs, reaching high accuracies. The work can be further extended with different data sources as well as different classes.

The model has been successfully exported and loaded to an Android app together with the Tensorflow library, giving the ability to perform real-time predictions. L2 regularization has already been implemented to prevent overfitting, but another solution would be to add dropout, which is one of the most effective and commonly used regularisation techniques for neural networks to prevent overfitting models.

## REFERENCES

[1] "WISDM Lab: Dataset." [Online]. Available: http://www.cis.fordham.edu/wisdm/dataset.php#actitracker. [Accessed: 05-Apr-2019].

[2] A. Graves, G. Wayne, and I. Danihelka, "Neural Turing Machines," pp. 1–26, 2014.

[3] D. Britz, "Recurrent Neural Networks Tutorial, Part 3 – Backpropagation Through Time and Vanishing Gradients – WildML," 2015. [Online]. Available: http://www.wildml.com/2015/10/recurrent-neural-networks-tutorial-part-3-backpropagation-through-time-and-vanishing-gradients/.

[4] "A Gentle Introduction to Exploding Gradients in Neural Networks." [Online]. Available: https://machinelearningmastery.com/exploding-gradients-in-neural-networks/. [Accessed: 07-Apr-2019].

[5] "The Vanishing Gradient Problem – Towards Data Science." [Online]. Available: https://towardsdatascience.com/the-vanishing-gradient-problem-69bf08b15484. [Accessed: 07-Apr-2019].

[6] D. N. T. How, K. S. M. Sahari, H. Yuhuang, and L. C. Kiong, "Multiple sequence behavior recognition on humanoid robot using long short-term memory (LSTM)," *2014 IEEE Int. Symp. Robot. Manuf. Autom. IEEE-ROMA2014*, no. November, pp. 109–114, 2015.

[7] "Create Segments and Labels." [Online]. Available: https://gist.github.com/ni79ls/6701d610b031c6e31e0fd43be1eafe4f.

[8] G. Hoffman, "GarrettHoffman/lstm-oreilly: How to build a Multilayered LSTM Network to infer Stock Market sentiment from social conversation using TensorFlow." [Online]. https://github.com/GarrettHoffman/lstm-oreilly.[Accessed:20-Apr-2019].

[9] "Tensorflow Placeholders - Databricks.com." [Online]. Available: https://databricks.com/tensorflow/placeholders. [Accessed: Apr-2019].

[10] D. P. Kingma and J. Ba, "Adam: A Method for Stochastic Optimization," pp. 1–15, 2014.

[11] "Differences between L1 and L2 as Loss Function and Regularization," 2013. [Online]. Available: http://www.chioka.in/differences-between-l1-and-l2-as-loss-function-and-regularization/. [Accessed: 07-Apr-2019].

[12] S. Sharma, "Epoch vs Batch Size vs Iterations – Towards Data Science," 2017. [Online]. Available: https://towardsdatascience.com/epoch-vs-iterations-vs-batch-size-4dfb9c7ce9c9. [Accessed: 14-Apr-2019].

[13] Pranesh Vallabh, et.al., "Fall detection monitoring systems: a comprehensive review", Journal of Ambient Intelligence and Humanized Computing, Vol.9, no.6, pp.1809-1833, 2019.

[14] Babedi Betty Letswamotse, et.al., "Software Defined Wireless Sensor Networks (SDWSN): A Review on Efficient Resources, Applications and Technologies", Journal of Interent Technology, Vol.19, no.5, pp.1303-1313, 2018.

[15] Babedi Betty Letswamotse, et.al., "Software defined wireless sensor networks and efficient congestion control", IET Networks, vol.7, no.6, pp.460-464, 2018.

[16] Johan Wannenburg, et. al., "Wireless Capacitive-Based ECG Sensing for Feature Extraction and Mobile Health Monitoring", IEEE Sensors Journal, Vol. 18, no.14, pp.6023-6032, 2018.

[17] Arun C. Jose, "Improving Home Automation Security; Integrating Device Fingerprinting into Smart Home", IEEE Access, Vol. 4, pp.5776-5787, 2016.

[18] Zhong-qin Wang, et. al. "Measuring the similarity of PML documents with RFID-based sensors", International Journal of Ad Hoc and Ubiquitous Computing, Vol.17, no.2, pp.174-185, 2014.